**Title**

"Volume Transfer": A New Design Concept for Fabric-Based Pneumatic Exosuits

*Chendong Liu, Dapeng Yang\*, Jiachen Chen, Yiming Dai, Li Jiang, Hong Liu*

C. Liu, D. Yang, J. Chen, Y. Dai, L. Jiang, H. Liu
State Key Laboratory of Robotics and System, Harbin Institute of Technology, 150001 Harbin, China
E-mail: yangdapeng@hit.edu.cn

((Abstract text. Maximum length 200 words. Written in the present tense.))
The fabric-based pneumatic exosuit is now a hot research topic because it is lighter and softer than traditional exoskeletons. Existing research focused more on the mechanical properties of the exosuit (e.g., torque and speed), but less on its wearability (e.g., appearance and comfort). This work presents a new design concept for fabric-based pneumatic exosuits: "Volume Transfer", which means transferring the volume of pneumatic actuators beyond the garment's profile to the inside. This allows for a concealed appearance and a larger stress area while maintaining adequate torques. In order to verify this concept, we develop a fabric-based pneumatic exosuit for knee extension assistance. Its profile is only 26mm and its stress area wraps around almost half of the leg. We use a mathematical model and simulation to determine the parameters of the exosuit, avoiding multiple iterations of the prototype. Experiment results show that the exosuit can generate a torque of 7.6Nm at a pressure of 90kPa and produce a significant reduction in the electromyography activity of the knee extensor muscles. We believe that "Volume Transfer" could be utilized prevalently in future fabric-based pneumatic exosuit designs to achieve a significant improvement in wearability.

## 1. Introduction

The exoskeleton is a kind of wearable mechanical assistive device with broad application prospects in the field of medical rehabilitation.[1-3] Traditional rigid exoskeletons are heavy, costly, lack flexibility, and have poor human-machine interaction performance.[4,5] In contrast, soft exoskeletons (exosuits) made with soft materials are lightweight, soft, low cost, and have better human- machine interaction performance.[6,7] In the past decade, two technological routes have been developed for exosuits: contraction exosuits and expansion exosuits.[8] Contraction exosuits use contraction actuators, most notably Bowden cables, as well as pneumatic artificial muscles, shape memory alloys, and so on.[9-11] The lower-limb exosuit designed by Harvard using Bowden cables actuation has been a huge success and has already yielded commercial applications.[12-22] However, since contraction exosuits use bionic contraction actuators with short force arms to actuate the joints, this will cause large shear forces on the human skin, which makes people uncomfortable.[8,23]

Expansion exosuits are actuated by soft pneumatic actuators, and the force is mainly perpendicular to the human skin, so the excessive shear force problem in contraction exosuits can be avoided to a certain extent.[8] There are two main types of soft pneumatic actuators used in expansion exosuits: silicone-based and fabric-based. Silicone-based pneumatic actuators, a type of actuator often employed in soft robots, have generated a great deal of development in the past decade. When it is used in the field of exosuits, the main application is in the upper limb, especially hand rehabilitation.[24,25] Fabric-based pneumatic actuators have been developed over the last five years,[26-31] and due to the stronger tensile properties of fabric, these actuators have a higher load carrying capacity than silicone-based soft pneumatic actuators.[32,33] Moreover, fabrics used as actuators can be easily attached to fabrics used as common garments using sewing, thus realizing a suit-liked exoskeleton design for excellent human-machine interaction.

Many researchers have made outstanding contributions to the study of fabric-based pneumatic exosuits. The works at Harvard University focused on the shoulder and elbow joints of the upper limb.[34-38] The research of Walsh et al. on the shoulder exosuit has been used in patients with amyotrophic lateral sclerosis. In addition to the upper limb, fabric-based pneumatic exosuits for the hip and knee joints of the lower limb have sprung up.[39-46] There is also the fabric-based pneumatic exosuit that targets the back muscles.[47]

The main concerns of current research on fabric-based pneumatic exosuits are how to design the structure to produce effective assistance for joints of the human body, [39-41,43,46] how to improve the assisting torque, [42] or how to improve the response speed.[44,45] To the best of our knowledge, there is no work discussing the following questions: how can fabric-based pneumatic exosuits be made to have a more flattering appearance so that they can be concealed within garments, avoiding users' sense of shame, and reducing the disturbance to daily life? How can the structure be designed to make it more comfortable for users? And these issues around wearability would have been exactly where the advantages of soft fabric-based actuation methods over traditional actuation methods lie, necessitating further in-depth research.

This work presents a new design concept for fabric-based pneumatic exosuits: "Volume Transfer", which means transferring the volume of pneumatic actuators beyond the garment's profile to the inside. This allows for a concealed appearance and a larger stress area while maintaining adequate torques. The concealed appearance means that the exosuit can be worn inside the garment, thus avoiding the users' sense of shame and reducing the disturbance to daily life. A larger stress area means less compression and users will feel more comfortable. In order to demonstrate the practical application of the "Volume Transfer" design concept, we developed a fabric-based pneumatic exosuit for knee extension assistance. It has a profile of about 26mm and can be concealed in common garments. Its stress area wraps around almost half of the leg. We used an analytical mathematical model based on the principle of virtual work and finite element simulation to determine the design parameters of the exosuit, avoiding multiple iterations of the prototype. During the design calculation process we used "Volume Transfer" to obtain a 37.3% reduction in the profile and a 149.7% increase in the stress area, with only a 0.34% decrease in the torque. The prototype was fabricated using heat sealing and sewing. The results of the mechanical experiments show that the exosuit can produce a torque of 7.6Nm at an air pressure of 90kPa. The surface electromyography (sEMG) experiments show that the exosuit can produce a significant reduction in the activity of leg muscles (rectus femoris, vastus lateralis, vastus medialis). Our research shows that the exosuit designed under the "Volume Transfer" design concept can realize a concealed appearance and increase the stress area, while maintaining sufficient actuation forces and a noticeable assistance effect.

## 2. Design Concept and Prototype

In order to demonstrate the practical application of the "Volume Transfer" design concept, we developed a fabric-based pneumatic exosuit for knee extension assistance, as shown in **Figure 1**a. This exosuit was made entirely from fabric and no rigid parts were used. Knee extension can be assisted by relying on the torque generated by the seven cylindrical fabric-based pneumatic actuators (blue in the figure) as they are bent.

Our design concept of "Volume Transfer" is inspired by the I-section steels (Figure 1b). The I-section is a very common form of beam sections in mechanical engineering. The reason for the superior bending resistance of the I-section over the rectangular section with the same area is that the I-section can actually be considered to be obtained by transferring the volume of the center portion of the rectangular section (cyan in the figure) to the edges. And according to the theory of material mechanics, the same volume of material arranged at the edges contributes more bending resistance compared to that arranged in the center. Going back to the design of exosuits, it is not difficult to achieve a concealed appearance for exosuits that use the contraction actuation method, [48-51] since the actuation force of most contraction actuators is not limited by the volume of the actuator. However, in the case of fabric-based pneumatic exosuits using pneumatic actuators, excessively reducing the volume of actuators in order to achieve a concealed appearance will bring about a significant reduction in actuation forces, weakening the assistance effect. There is a paradox here in that it is difficult to have both the appearance and the actuation force. This paradox can be solved by using the design concept: "Volume Transfer".

Figure 1c shows the view of the user wearing the exosuit. It can be seen that the exosuit we developed wraps around almost the entire back part of the leg profile. Such a large stress area will result in a more comfortable wearing experience. The cross-section along the radial direction of the leg visualizes the "Volume Transfer" design concept. The cylindrical fabric-based pneumatic actuator we used is a common form of actuators used in fabric-based pneumatic exosuits now. [52] Following the conventional arrangement, two actuators are arranged in the leg (black dashed lines). [44] This is where the paradox of the appearance versus the actuation described above arises. Actuators cannot be concealed if their volume exceeds the profile line of the garment (green solid line). If the volume of the actuators is reduced to within the profile line of the garment, the actuation force will be greatly reduced. How can we reduce the volume of the actuators and still maintain sufficient actuation forces? Our approach is to "transfer" the volume, rather than to reduce it. By transferring (orange arrows) the volume (cyan) of the actuators beyond the profile line of the garment to the inside, making full use of the space inside the garment's profile to arrange the actuators, the paradox between the appearance and the actuation, which is difficult to be obtained at the same time, can be solved. That's the design concept of "Volume Transfer".

The result of "Volume Transfer" is that a single, larger actuator is replaced by multiple, smaller actuators (blue solid lines). "Volume Transfer" also brings another advantage, i.e., an increase in the stress area (the black thick dashed line in the figure denotes the stress area before "Volume Transfer", and the blue thick solid line denotes the stress area after "Volume Transfer"). Figure 1d shows the concealed appearance of the exosuit. The prototype developed in this work has a profile of 26 mm (the distance between the skin surface of the leg and the profile line of the garment) and can be concealed in common garments.

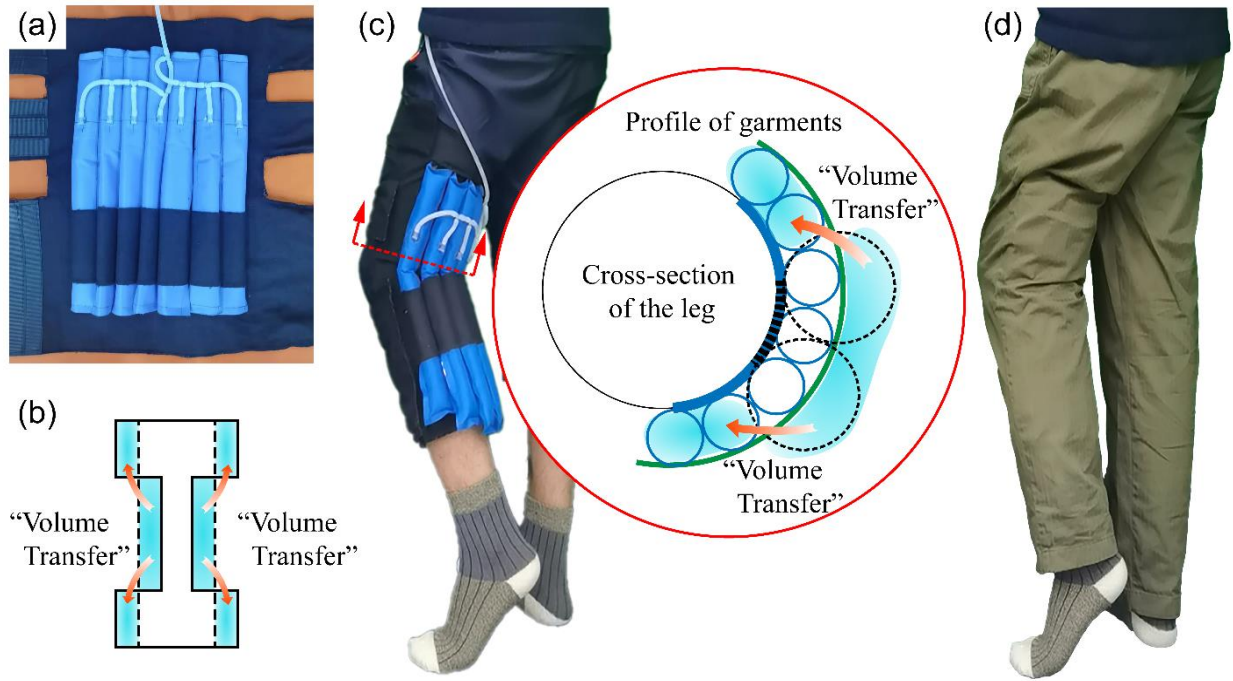

**Figure 1.** Design concept and prototype. a) The fabric-based pneumatic exosuit fabricated for the demonstration of "Volume Transfer" is actuated with seven cylindrical fabric-based pneumatic actuators. b) The "Volume Transfer" concept in the design of I-section steels. The cyan area denotes the area of "Volume Transfer". c) A user wearing the exosuit. The cross-section of the leg visualizes the "Volume Transfer" design concept. The black solid line denotes the profile of the leg. The green solid line denotes the profile of garments. The black dashed line denotes actuators before "Volume Transfer". The blue solid line denotes actuators after "Volume Transfer". The orange arrows denote the direction of the "Volume Transfer". The black thick dashed line denotes the stress area before "Volume Transfer". The blue thick solid line denotes the stress area after "Volume Transfer". d) The concealed appearance of the exosuit.

## 3. Design Calculation Process of the Exosuit

Due to the difficulty in accurate modeling of soft robots, most of the design of soft robots depends on the trial production of prototypes, which requires several iterations to get a satisfactory design. In this work, a simple but effective method was used to design the key parameters of the exosuit. Although this method does not accurately describe the behavior of the designed exosuit, designing an exosuit under the guidance of this method avoids the need for multiple rounds of prototype iterations, and an exosuit that meets the design indicators can be obtained in only one trial run.

As mentioned earlier, the design of the exosuit proposed in this work requires the use of multiple cylindrical fabric-based pneumatic actuators, so the mechanical behavior of a single actuator was first investigated. As shown in **Figure 2**a, three parameters are required to describe the bending behavior of a single actuator: the bottom area $S$ of the cylinder (gray area in the figure), the diameter $d$ of the cylinder, and the bending angle $\theta$ of the cylinder. The shaded area denotes the region where the cylindrical actuator interferes in space due to bending, and its volume is denoted as $V$. The root cause of the cylindrical actuator's ability to generate torques during bending lies in this part of the volume that interferes in space. A core mathematical model describing the mechanical behavior of fabric-based pneumatic actuators has been proposed, [52] and its core idea is based on the principle of virtual work. The expression for the work done by the expansion of the gas inside the actuator, $W_1$, is given by Equation (1):

$$W_1 = \int p dV \tag{1}$$

$p$ denotes the air pressure inside the actuator, and the expression for the work done by the output torque $W_2$ is given by Equation (2):

$$W_2 = \int \tau d\theta \tag{2}$$

$\tau$ denotes the output torque of a single actuator, and according to the principle of virtual work, these two parts of the work done should be equal (Equation (3)):

$$W_1 = W_2 \tag{3}$$

An expression for the output torque $\tau$ of a single actuator can thus be obtained as:

$$\tau = p \frac{dV}{d\theta} \tag{4}$$

The exosuit we designed consists of multiple cylindrical actuators wrapped around the leg, and their radial cross-sections are shown in Figure 2b, $n$ denotes the number of cylindrical actuators, $D$ denotes the diameter of the leg cross-section (here the leg cross-section is approximated as a circle), and $\alpha$ denotes the angle of the lines from the center of the cross-section to the adjacent cylindrical actuators. We consider the output torque of an exosuit including multiple cylindrical actuators as a superposition of the single cylindrical actuator, then the output torque $T$ of the exosuit can be expressed as:

$$T = np \frac{dV}{d\theta} \tag{5}$$

Equation (5) assumes that the bending angle of each actuator is the same $\theta$, equal to the bending angle of the knee joint of the leg. This is actually inaccurate because the actuator is not attached to the leg via a rigid connector. Our exosuit does not include any rigid pieces and is attached entirely with fabric. This allows for a more comfortable wearable experience, but due to the flexibility of the fabric, we cannot constrain the bending angle of each actuator to be strictly equal to the bending angle of the leg's knee joint. Trying to describe this angular inconsistency in analytic mathematical formulas is very difficult. Fortunately the error introduced by the calculation using Equation (5) is not unacceptable, as evidenced by the experimental test results of the prototype we finally designed.

The question now becomes how to calculate the volume of the interfering part $V$. For a cylindrical actuator, the expression for $V$ is given by Equation (6):

$$V = Sd \tan(\frac{\theta}{2}) \tag{6}$$

The relationship between the bottom area $S$ of the cylinder and the diameter $d$ of the cylinder is given by Equation (7):

$$S = \frac{\pi d^2}{4} \tag{7}$$

Thus, we obtain the Equation (8) for calculating the output torque $T$ of the exosuit composed of multiple cylindrical actuators:

$$T = \frac{\pi n p d^3}{8\cos^2(\frac{\theta}{2})} \tag{8}$$

The significance of Equation (8) lies in obtaining the relationship between the exosuit output torque $T$ and four parameters, namely, the internal pressure $p$ of the actuator, the actuator bending angle $\theta$, the number of actuators $n$, and the diameter $d$ of the actuator. $T$, $p$, and $\theta$ is determined by the demand for using. These three parameters are design indicators, while $n$ and $d$ is the real parameters we want to design. That is, we have to decide how many actuators and what diameter to use in order to generate enough output torques with a limited amount of air pressure and a specific actuator bending angle.

The design calculation graph (Figure 2d) was used to determine the number of actuators $n$, and the diameter $d$. We hope that our final design of the exosuit will be able to produce a torque of no less than 6Nm at a bending angle of 90° and an air pressure of 90kPa (There are many benefits to using a low air pressure, such as increased safety, prevention of injuries to the user, increased reliability and service life of actuators, prevention of leaks, and reduced performance demands on the pneumatic system, resulting in a lighter pneumatic system.), and this torque does not decrease to less than 1Nm when the bending angle of the actuator decreases to 20°.

Two design calculation graphs corresponding to the coordinates (90°, 90 kPa) and (20°, 90 kPa) in Figure 2d can be obtained by using Equation (8) (In fact, any coordinate value in Figure 2d corresponds to a design calculation graph and which one to use depends entirely on the designer's design indicators). Using the contour lines in the design calculation graphs, we can visualize the output torque that can be produced by a given actuator diameter $d$ and a number of actuators $n$, thus guiding the selection of parameters.

It should be noted that parameters $n$ and $d$ are not independent of each other. As shown in Figure 2b, the actuators must be arranged under the coronal plane of the leg so that the force generated by the actuators is in the form of pressure on the surface of the leg, thus assisting in knee extension. Actuators arranged above the coronal plane of the leg will create tension on the leg binding fabric, which is not our design intent. For the design of this work, the diameter at the knee cross-section is taken as an approximation of the diameter of the leg, i.e., the value of $D$ is taken as 120 mm. Assuming that the actuator cross-section circle and the leg cross-section circle are tangent, and the neighboring actuator cross-section circles are tangent to each other, the relationship between the leg diameter $D$, the actuator diameter $d$, and the angle $\alpha$ of the lines from the center of the actuator cross-section circle to the center of the leg cross-section circle can be obtained according to the Cosine Theorem as Equation (9):

$$\cos\alpha = \left(\frac{2(\frac{D+d}{2})^2 - d^2}{2(\frac{D+d}{2})^2}\right) \tag{9}$$

To ensure that all actuators are arranged under the coronal plane of the leg, the number of actuators $n$ (which can only be taken as an integer) should satisfy Equation (10):

$$n < [\frac{\pi}{\alpha}] + 1 \tag{10}$$

Plotting Equation (9) and Equation (10) in the design calculation graph of Figure 2d, we obtain the black solid line that descends in a step-like fashion, and the number of actuators $n$ should lie below this line.

In order to better realize our idea of concealing the exosuit inside a common garment, the actuator diameter $d$ should be taken to be as small as possible while still meeting the other indicators. If we follow the conventional design method in the past, we can choose the triangular point in Figure 2d as our design point ($d = 38$mm, $n = 2$). But we use the new design concept: "Volume Transfer" to transfer (orange arrow) the design point to the star point ($d = 25$mm, $n = 7$). This results in a 37.3% reduction in the exosuit profile, a 149.7% increase in the stress area, but only a 0.34% decrease in the torque. That is the great power of "Volume Transfer".

The design calculation graph at the coordinates (20°, 90kPa) was used for verification. The torque generated by the design point ($d = 25$mm, $n = 7$). was above 1Nm, meeting the design indicators.

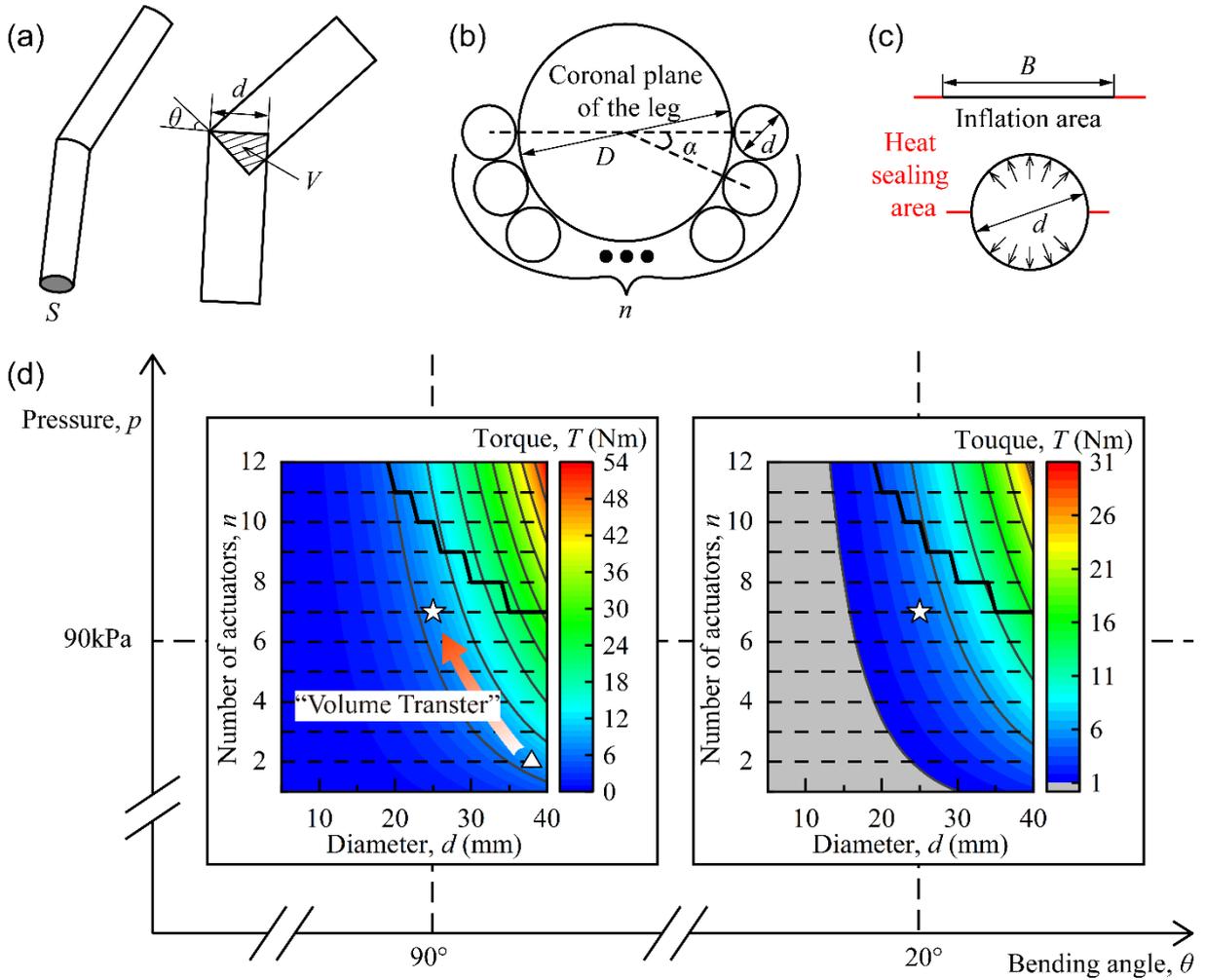

**Figure 2.** The parameters of the exosuit and the design calculation graphs. a) Schematic diagram of the shape and cross-section of a cylindrical actuator when it is bent. $S$ denotes the area of the bottom of the cylinder. $\theta$ denotes the angle at which the cylindrical actuator is bent. $d$ denotes the diameter of the cylinder. The shaded portion denotes the region of spatial interference due to the bending of the cylindrical actuator, and its volume is denoted as $V$. b) Schematic diagram of the leg cross-section. $n$ denotes the number of actuators. $D$ denotes the diameter of the leg cross-section. $\alpha$ denotes the angle of the lines connecting the centers of adjacent cylindrical actuator cross-sections to the center of the leg cross-section. All actuators are arranged under the coronal plane of the leg. c) Schematic diagram of the changes before and after inflation in an actuator made of two pieces of fabric by heat sealing. d) Design calculation graphs at different bending angles and inflation pressures. The black stepped solid line denotes the maximum value of $n$. The orange arrow denotes the direction of "Volume Transfer". The triangular point denotes the conventional design point before "Volume Transfer". The star point denotes the new design point obtained after "Volume Transfer".

But determining the diameter $d$ after the actuator has inflated into a cylindrical shape does not mean the end of the design, it has to be translated into the width $B$ before inflation (Figure 2c) in order to facilitate fabric cutting. Assuming that the sealing fabric used to fabricate the actuator is completely non-stretchable, the diameter $d$ after inflation and the width $B$ before inflation approximately satisfy Equation (11):

$$2B = \pi d \tag{11}$$

Then the value of $B$ corresponding to $d = 25$mm is 39.3mm. The value of $B$ was rounded upwards to 40mm for ease to cut. However, considering that there is actually some degree of stretching of the sealing fabric after inflation, this may cause the actual value of $d$ to be larger than the design value. To assess the magnitude of this error, we used the method of finite element simulation. The sealing fabric we used to fabricate the cylindrical fabric-based pneumatic actuator is N210D nylon fabric, whose mechanical properties have been tested in detail. [53] We quoted the results of their tests, as shown in **Table 1**:

**Table 1.** Properties of N210D TPU nylon fabrics.

| Parameters | Value |
| --- | --- |
| Density [kg·m$^{-3}$] | 1375 |
| Thickness [mm] | 0.2 |
| Elastic modulus $E_1$ [MPa] | 343.1 |
| Elastic modulus $E_2$ [MPa] | 134.6 |
| Poisson ratio $u_{12}$ | 0.412 |
| Poisson ratio $u_{21}$ | 0.326 |
| Shear modulus $G_{12}$ [MPa] | 14.8 |

We used abaqus/explicit to simulate the deformation of the actuator. We processed the fabric sheets as homogeneous continuum shells made of anisotropic lamina materia. Inflation pressure was simulated using homogeneous pressure. The simulation results are shown in **Figure 3**. The diameter of the actuator $d$ after inflation is about 28.8 mm. There is a 15% enlargement compared to the design value, but it still conforms to the design calculation graph in Figure 2d. So we carried out the trial production of a single cylindrical fabric-based pneumatic actuator according to $B = 40$mm and actually measured its diameter (Figure 3), which was about 26mm, with only 4% error compared to the design value. After the above design calculation process, we determined the important parameters for the exosuit design, i.e., the number of actuators *n* is 7, the diameter *d* is 25mm, and the width *B* is 40mm, thus completing the whole design process.

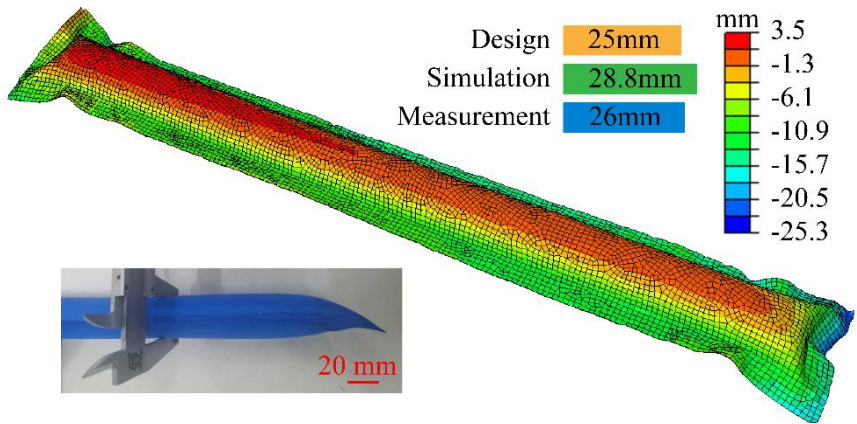

**Figure 3.** Finite element simulation and experimental measurements of a cylindrical fabric-based pneumatic actuator. For an actuator with a width of *B* = 40mm, the simulation value of the diameter *d* after inflation is 28.8mm, and the measurement value is 26mm, compared to the design value 25mm.

## 4. Fabrication of the Exosuit

To fabricate the exosuit, two types of fabrics were used: a single-sided TPU nylon fabric (N210D, Jiaxing Yingcheng, China) and a common denim fabric. Single-sided TPU nylon fabric is easy to seal by heat pressing, and we use it as the sealing fabric for the actuator. Denim fabrics are common clothing fabrics that are comfortable to wear and can be attached to sealing fabrics by sewing. The method of fabricating a single actuator is shown in **Figure 4**a. We used an iron (200°C) to press all four sides of the sealing fabric cut into rectangles, with a width of about 1cm in the heat sealing area. Pagoda-type air nozzles (Beiloquan, China) were glued (Xunlei, China) to allow the actuator to be connected to a silicone hose (Biling, China). Seven identical actuators were fabricated using this method. A sewing machine was then used to sew the actuators together with the fixing fabric (denim) and the binding fabric shown in Figure 4b. When sewing, it is important that the sewing threads are restricted within the heat sealing area to avoid air leaks. The purpose of the fixing fabric is to secure the actuators reliably to the binding fabric and avoid undesired slippage. Velcro was sewn to the ends of the binding fabric so that the entire device can be worn comfortably and reliably on the leg.

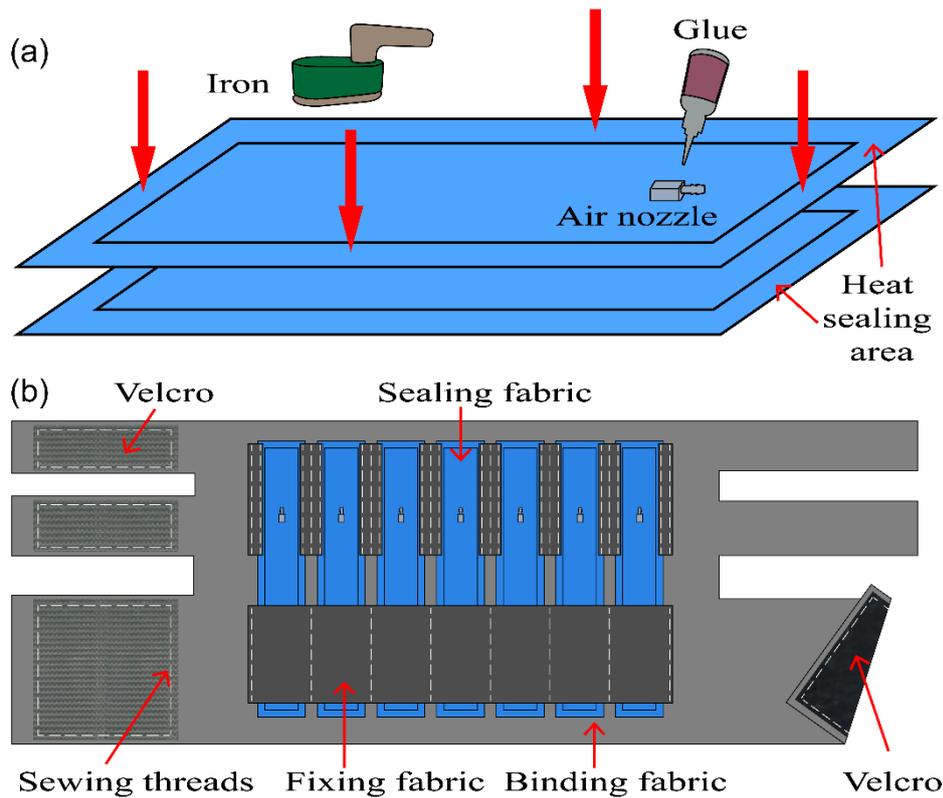

**Figure 4.** Fabrication process of the exosuit. a) Fabrication process for a single actuator. The single-sided TPU nylon fabric was used as the sealing material. The actuator was sealed by heat pressing around the edges, and the air nozzle was bonded using glue. b) Fabrication process for an entire exosuit. The method in a) was used to fabricate seven actuators. These actuators were attached to the fixing and binding fabrics by sewing. Velcro was used to strap the exosuit to the leg.

## 5. Mechanical Experiment

The mechanical properties of the developed exosuit were tested with the aim of obtaining the output torque of the exosuit at different air pressures and bending angles. The experimental equipment we used is shown in **Figure 5**a. The table was made of aluminum profiles, so that the experimental equipments can be easily fixed, and the position can be flexibly adjusted. The exosuit was strapped to the rotating beam with aluminum profiles as the backbone. The rotating beam can be rotated in the plane over a range of 180°. A protractor was used to measure the angle of the rotating beam. This angle can be approximated as the bending angle of the exosuit. A force sensor including an accompanying digital meter (DYLF-102, DAYSENSOR, China) was used to measure the force at the end of the rotating beam, which was multiplied by the force arm measured before the experiment to give the exosuit's output torque. An air compressor (ED-0204, Eidolon, China) (not shown in the figure due to its large size) was connected to the red air hose and supplied air to the exosuit through a pressure reducing valve (IR2010-02-A, SMC, Japan) which allows for manual adjustment of the pressure and a manual switch valve (Juoji, China) which allows for easy control of the start and end of the experiment. The air compressor, pressure reducing valve, force sensor and digital meter used in the experiment were powered by AC and DC power sources.

Figure 5b shows the experimentally measured output torques of the exosuit at different air pressures (10~90kPa) and different angles (20°~90°). It can be intuitively seen from the figure that the output torque of exosuit is positively correlated with the air pressure and the bending angle. At the air pressure of 90kPa and the bending angle of 90°, the maximum output torque of the exosuit is 7.6Nm, which exceeds 6Nm. At the air pressure of 90kPa and the bending angle of 20°, the output torque is 2.2Nm, which exceeds 1Nm. The experimental results show that the output torque of the exosuit meets our pre-determined design indicators, thus demonstrating that an exosuit that meets the usage requirements can be efficiently designed using the design calculation method we described earlier. Substituting the measured diameter of the cylindrical fabric-based pneumatic actuator (26mm) into the design calculation graph in Figure 2d, we obtained that the theoretical calculated value of the exosuit's maximum output torque is 8.7 Nm, which exists an error of 14.5% compared to the actual value. This error is mainly caused by the idealized Equation (5), and it is good to know that this error is not enough to affect our design. The test results in Figure 5b can be drawn in the two-dimensional form (Figure 5c and Figure 5d), so that the relationship between the output torque, the air pressure and the bending angle can be more clearly observed. As shown in Figure 5c, the relationship between the output torque and the air pressure at a particular angle is approximately linear, and the rate of change increases essentially with the bending angle. It is worth noting that the output torque at angles of 60° and 70° is surprisingly smaller than that at the angle of 50°, which is not in line with theoretical expectations. We believe the cause of this phenomenon is that the binding fabric we designed creates a large amount of slack in this angular range, which results in an abnormal decay of the exosuit's output torque.

As shown in Figure 5d, there is not a linear relationship between the torque and the bending angle as there is between the torque and the air pressure, but only a positive correlation. The torque decay phenomenon described above due to binding slack can also be observed at the bending angle of 60°. In conclusion, the mechanical properties of the exosuit are slightly lower than the theoretical design values, but still meet the requirements for use.

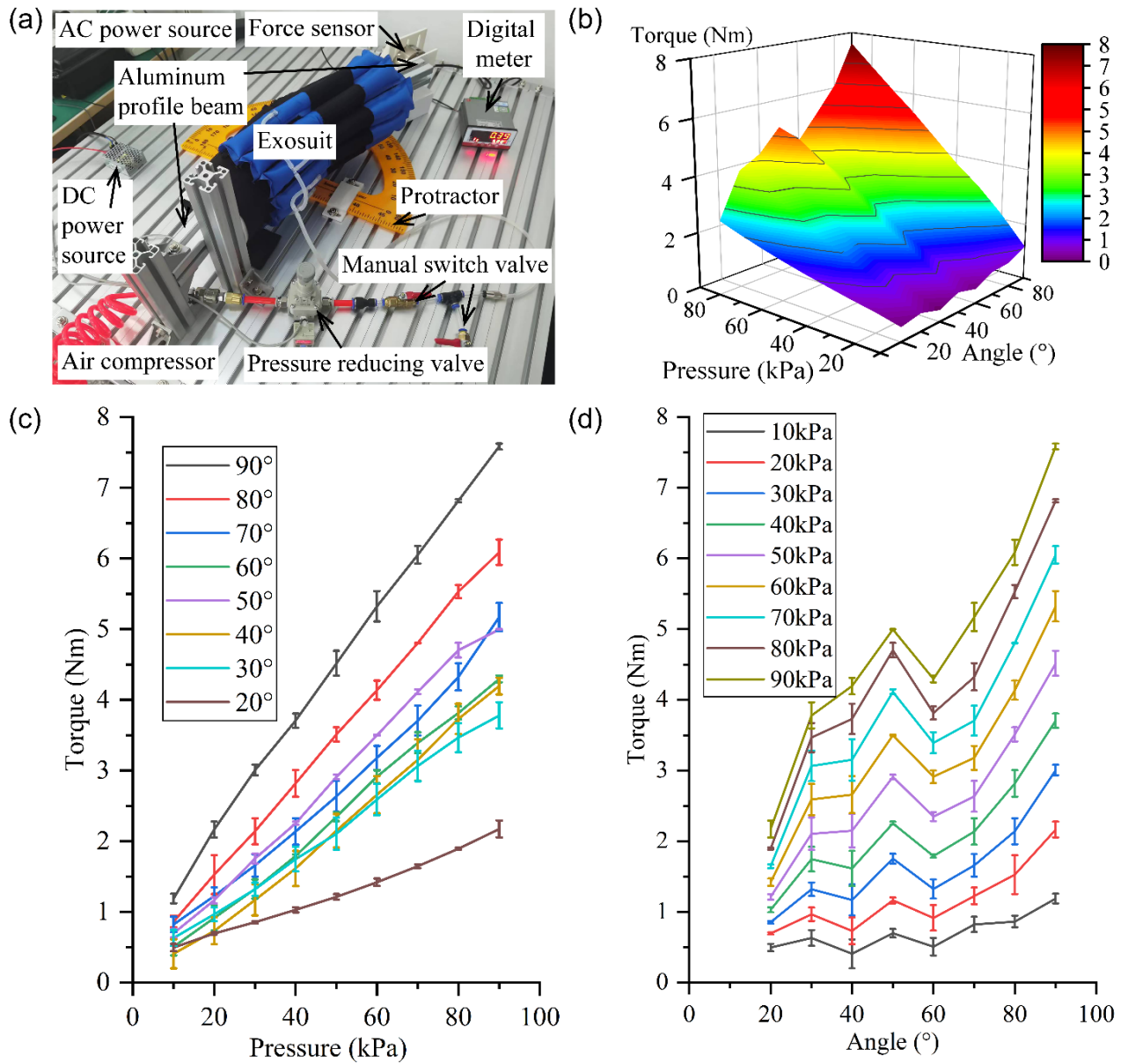

**Figure 5.** Mechanical experiment of the exosuit. a) Mechanical testing experimental platform. b) Output torques at different air pressures and angles. c) Relationship between output torques and air pressures. d) Relationship between output torques and angles.

## 6. Surface Electromyography Experiment

We evaluated the actual assisting effect of our exosuit on knee extension by the surface electromyography (sEMG) experiment. Five participants were recruited for the experiment. Information on the five participants is shown in **Table 2**. For each participant, a description about the purpose and nature of the experiments was provided. Our experiment was approved by the university's ethics committee and conformed to the declaration of Helsinki.

**Table 2.** Information on the participants.

| Number | Gender | Age | Height [cm] | Weight [kg] |
| --- | --- | --- | --- | --- |
| 1 | Male | 28 | 175 | 75 |
| 2 | Male | 27 | 178 | 70 |
| 3 | Male | 23 | 178 | 75 |
| 4 | Male | 29 | 170 | 80 |
| 5 | Male | 26 | 180 | 70 |

In the experiment, the participant wore the exosuit on the left leg with the body lying flat on a mat, and performed the yoga poses shown in **Figure 6**a: the thigh was lifted and kept perpendicular to the body, and the calf was kept horizontal. The leg muscles that drive knee extension are mainly rectus femoris, vastus lateralis, and vastus medialis, so we arranged surface electromyographic signal electrodes (Delsys, USA) on these three muscles, and the positions of the electrodes are shown in Figure 6b. When the participant performs the maneuver shown in Figure 6a, the three muscles mentioned above will fire to keep the calf horizontal without sagging under the force of gravity. If our exosuit is able to create effective assistance to knee extension, it will hold up the calf, thus relieving the burden on the three muscles mentioned above. So we asked participants to hold this posture for one minute without exosuit activation and then for one minute with exosuit activation. By comparing the strength of the sEMG signals of the participants in two one-minute periods, the assisting effect of our exosuit can be quantitatively assessed.

The results of sEMG signal testing are shown in Figure 6c (the signal was processed by 4th order Butterworth filtering from 10 Hz to 400 Hz). The test results show very significant differences between different participants. Our exosuit demonstrated almost perfect results in participant 1. With the help of exosuit, the sEMG signals of rectus femoris, vastus lateralis, and vastus medialis in participant 1 were reduced by 79.6%, 76.1%, and 83.5%, respectively, with an average of around 80%. But performance in other participants is less than perfect. Assistance in participant 2 is slightly less effective than participant 1, with average sEMG signal reductions in the range of 40% to 60%. Assistance in participant 3 and participant 4 shows a further decrease, with average sEMG signal reductions in the range of 10% to 40%. Participant 5 has the worst assistance effect, and his sEMG signals for vastus medialis even rose by 0.1% after the exosuit was used.

There are two main reasons why we believe there is such a huge difference. First, the stature of different participants varied. The dimensions of the exosuit were designed to fit the body of participant 1, so the exosuit could achieve a good bondage on participant 1. The other participants differs somewhat in stature from participant 1. The greatest difference is in participant 5, who was thinner, which results in some degree of binding slack. Second, the level of proficiency in the use of the exosuit varied across participants. Participant 1 had practiced a certain number of times before the experiment, but the other participants had not practiced in advance. This suggests that both customized sizing and adequate practice can help improve the exosuit's assistance.

The five participants who took part in the experiment were asked how they felt about using the exosuit. They are all satisfied with the comfort of the exosuit. One of the participants said: "I feel an even force holding my calf as if it is a large hand!" In conclusion, we demonstrate through sEMG experiments that our designed exosuit is able to produce significant assistance on knee extension, and this assistance varies significantly among individuals.

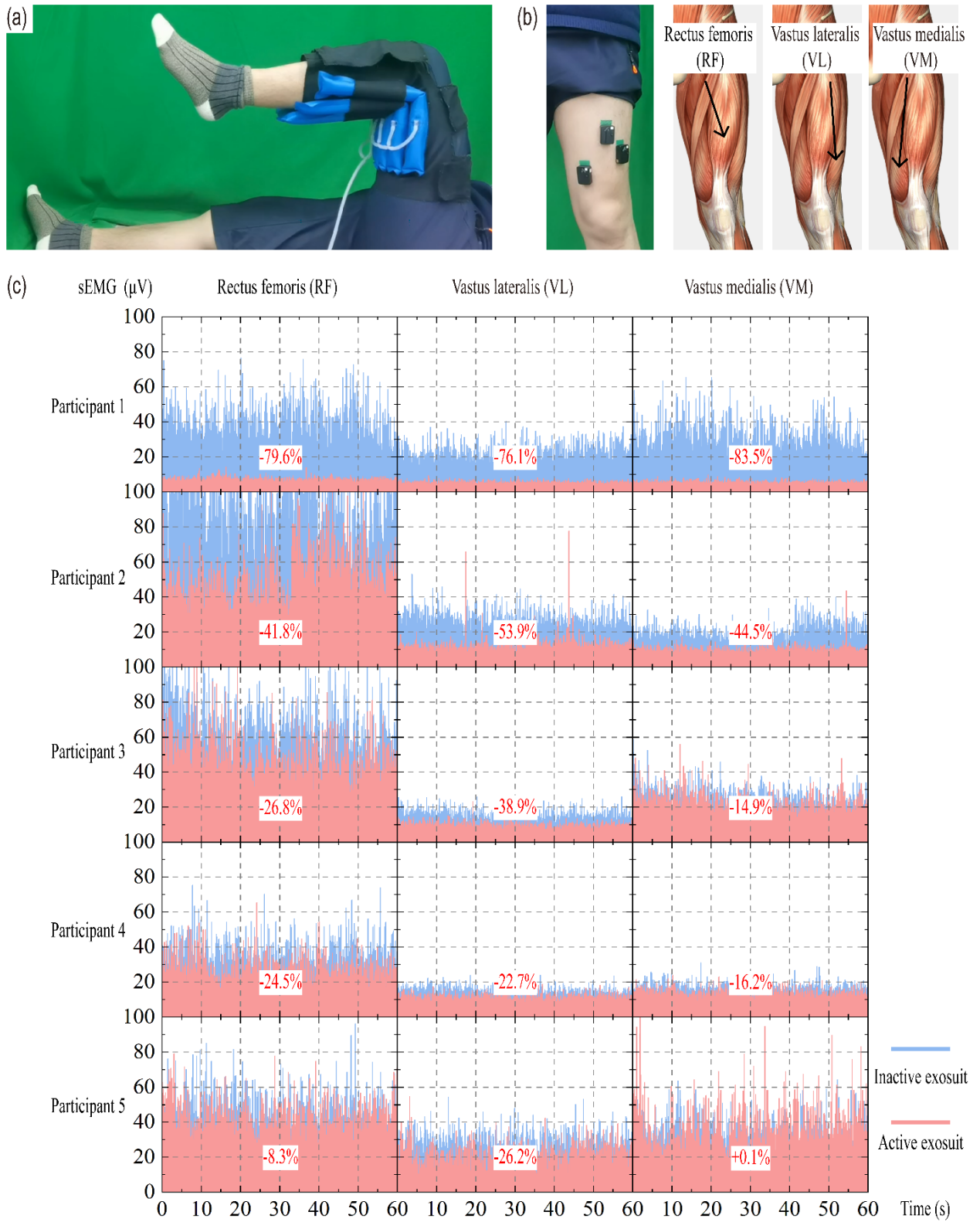

**Figure 6.** Surface electromyography experiment. a) The yoga poses performed by the participant. The participant lays flat on a mat with the left thigh raised, keeping it perpendicular to the body and the calf horizontal. b) Positions of surface electromyographic signal electrodes (Delsys, USA). Three electrodes were arranged in rectus femoris, vastus lateralis, and vastus

medialis. c) sEMG signals and their differences in three leg muscles and five participants with and without exosuit activation.

## 7. Conclusion

This work presents a new design concept for fabric-based pneumatic exosuits: "Volume Transfer", which means transferring the volume of the pneumatic actuators beyond the garment's profile to the inside. This allows for a concealed appearance and a larger stress area while maintaining adequate torques. The concealed appearance means that the exosuit can be worn inside the garment, thus avoiding the user's sense of shame and reducing the disturbance to daily life. A larger stress area means less compression and users will feel more comfortable. In order to demonstrate the practical application of the "Volume Transfer" design concept, we developed a fabric-based pneumatic exosuit for knee extension assistance. It has a profile of about 26mm and can be concealed in common garments. Its stress area wraps around almost half of the leg. We used an analytical mathematical model based on the principle of virtual work and finite element simulation to determine the design parameters of the exosuit, avoiding multiple iterations of the prototype. During the design calculation process we used "Volume Transfer" to obtain a 37.3% reduction in the profile and a 149.7% increase in the stress area, with only a 0.34% decrease in the torque. The prototype was fabricated using heat sealing and sewing. The results of the mechanical experiment show that the exosuit can produce a torque of 7.6Nm at an air pressure of 90kPa. The error of the measured value of the torque compared to the theoretical value is 14.5%. It can be observed in the mechanical experiments that the relaxation of the binding fabric leads to a decrease in the output torque of the exosuit. Surface electromyography experiments show that the exosuit produces a significant reduction in leg muscle (rectus femoris, vastus lateralis, vastus medialis) activities. Significant differences in exosuit's assistance can be observed in different individuals. Our research shows that the exosuit designed under the "Volume Transfer" design concept can realize a concealed appearance and increase the stress area, while maintaining sufficient actuation forces and a noticeable assistance effect.

Making exoskeletons to be apparels is an amazing trend in exoskeleton research. The Fabric-based pneumatic exosuit is currently the most promising technology for this. To make exoskeletons to be apparels, researchers should realize that the wearability is an important and worthwhile indicator, not just an optional "freebie". We believe that "Volume Transfer" could be utilized prevalently in the future fabric-based pneumatic exosuit design to achieve a significant improvement in wearability.

**Acknowledgements**

This work was supported by the National Natural Science Foundation of China (Grant No. 52075114) and State Key Laboratory of Robotics and System (HIT) (Grant No.SKLRS-2022-ZM-11).